\documentclass[10pt,twocolumn,letterpaper]{article}

\usepackage{cvpr}              

\usepackage{graphicx}
\usepackage{amsmath}
\usepackage{amssymb}
\usepackage{booktabs}
\usepackage[accsupp]{axessibility} 

\usepackage{multirow, makecell}
\usepackage{colortbl}  
\usepackage{float}
\usepackage{xcolor}
\usepackage{array}
\definecolor{Gray}{gray}{0.94}
\definecolor{LightCyan}{rgb}{0.88,1,1}

\newlength\savewidth\newcommand\shline{\noalign{\global\savewidth\arrayrulewidth
  \global\arrayrulewidth 1pt}\hline\noalign{\global\arrayrulewidth\savewidth}}
\newcommand{\tablestyle}[2]{\setlength{\tabcolsep}{#1}\renewcommand{\arraystretch}{#2}\centering\footnotesize}

\usepackage[pagebackref,breaklinks,colorlinks]{hyperref}

\usepackage[capitalize]{cleveref}
\crefname{section}{Sec.}{Secs.}
\Crefname{section}{Section}{Sections}
\Crefname{table}{Table}{Tables}
\crefname{table}{Tab.}{Tabs.}

\def\cvprPaperID{299} 
\def\confName{CVPR}
\def\confYear{2023}

\begin{document}

\title{
MoLo: Motion-augmented Long-short Contrastive Learning for \\ Few-shot Action Recognition
}

\author{
   \hspace{-0.4cm} Xiang Wang$^{1,2*}$
     \hspace{0.01cm} 
    Shiwei Zhang$^{2\dag}$
     \hspace{0.01cm} 
    Zhiwu Qing$^{1}$
     \hspace{0.01cm} 
    Changxin Gao$^1$ 
     \hspace{0.01cm}
   Yingya Zhang$^2$ 
     \hspace{0.01cm}
    Deli Zhao$^2$
    \hspace{0.01cm}
     Nong Sang$^{1\dag}$\\
    $^1$Key Laboratory of Image Processing and Intelligent Control,\\  \hspace{-0.5cm} School of Artificial Intelligence and Automation, Huazhong University of Science and Technology\\
     \hspace{-0.5cm}  $^2$Alibaba Group \\
{\tt\footnotesize \{wxiang,qzw,cgao,nsang\}@hust.edu.cn, \{zhangjin.zsw,yingya.zyy\}@alibaba-inc.com, zhaodeli@gmail.com}
}
\maketitle
\let\thefootnote\relax\footnotetext{$*$ Intern at Alibaba DAMO Academy. \hspace{1mm} $\dag$ Corresponding authors. }
\begin{abstract}
Current state-of-the-art approaches for few-shot action recognition achieve promising performance by conducting frame-level matching on learned visual features.
%
%
However, they generally suffer from two limitations:
i) the matching procedure between local frames tends to be inaccurate due to the lack of guidance to force long-range temporal perception;
%
ii) explicit motion learning is usually ignored, leading to partial information loss.
%
%
To address these issues, we develop a \textbf{Mo}tion-augmented \textbf{Lo}ng-short Contrastive Learning (MoLo) method that contains two crucial components, including a long-short contrastive objective and a motion autodecoder.
%
%
Specifically, the long-short contrastive objective is to
endow local frame features with long-form temporal awareness by maximizing their agreement with the global token of videos belonging to the same class.
The motion autodecoder is a lightweight architecture to reconstruct pixel motions from the differential features, which explicitly embeds the network with motion dynamics.
%
%
By this means, MoLo can simultaneously learn long-range temporal context and motion cues for comprehensive few-shot matching.
%
%
%
To demonstrate the effectiveness, we evaluate MoLo on five standard benchmarks, and the results show that MoLo favorably outperforms recent advanced methods.
%
The source code is available at \url{https://github.com/alibaba-mmai-research/MoLo}.
\end{abstract}

\section{Introduction}
\label{sec:intro}
Recently, action recognition has achieved remarkable progress and shown broad prospects in many application fields~\cite{TSN,TSM,Kinetics,EPIC-100,vivit}.
Despite this, these successes rely heavily on large amounts of manual data annotation,
which greatly limits the scalability to unseen categories due to the high cost of acquiring large-scale labeled samples.
To alleviate the reliance on massive data, few-shot action recognition~\cite{CMN} is a promising direction, aiming to identify novel classes with extremely limited labeled videos.
%
\begin{figure}[t]
  \centering
   \includegraphics[width=0.95\linewidth]{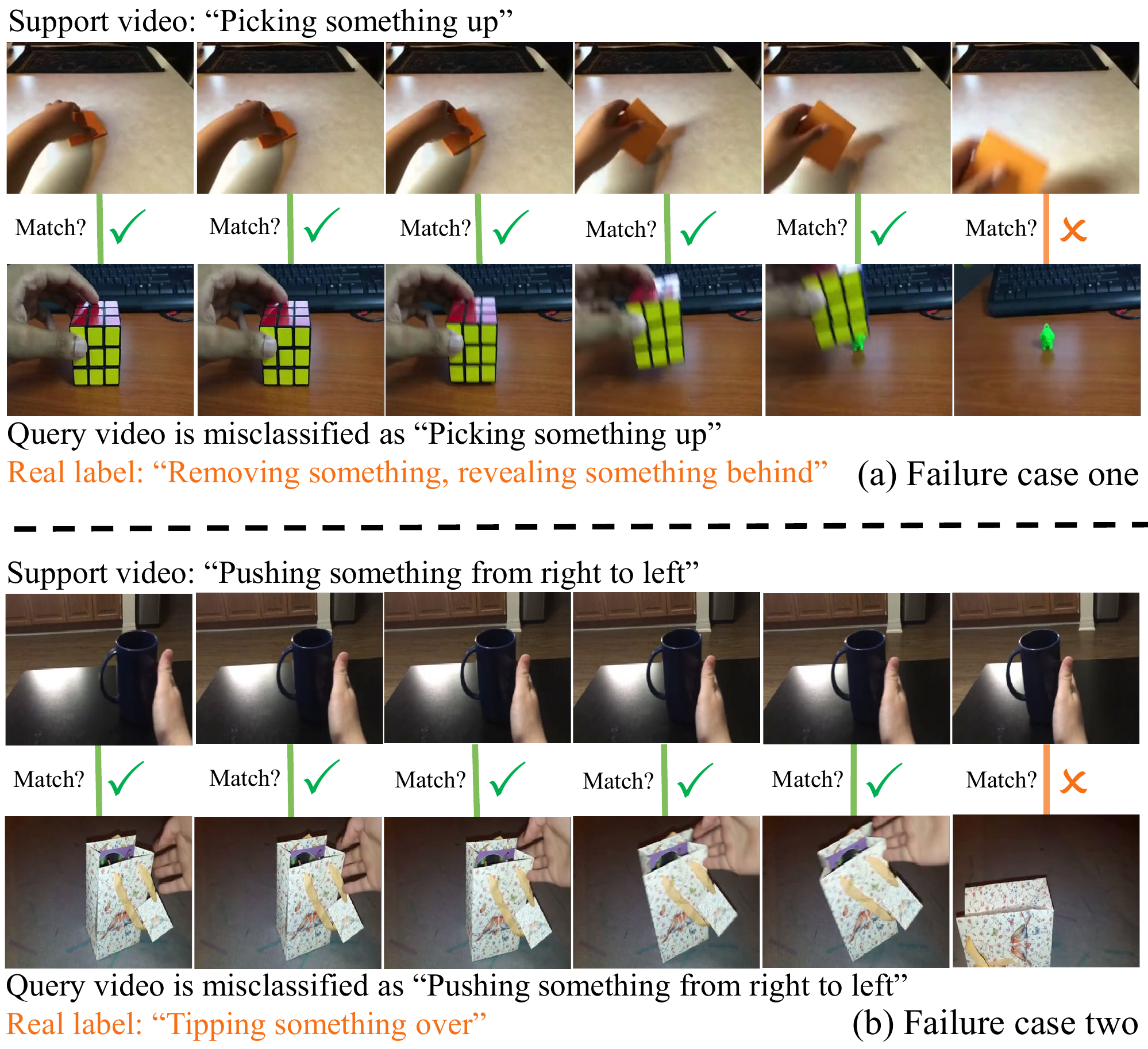}
\vspace{-3mm}
   \caption{Illustration of our motivation. 
   We show that most existing metric-based local frame matching methods, such as OTAM~\cite{OTAM}, can be easily perturbed by some similar co-existing video frames due to the lack of forced global context awareness during the support-query temporal alignment process.
   Example videos come from the commonly used SSv2 dataset~\cite{SSV2}.
   }
   \label{fig:Motivation}
   \vspace{-4mm}
\end{figure}

Most mainstream few-shot action recognition approaches~\cite{OTAM,TRX,HyRSM,huang2022compound} adopt the metric-based meta-learning strategy~\cite{MatchNet} that learns to map videos into an appropriate feature space and then performs alignment metrics to predict query labels.
Typically, OTAM~\cite{OTAM} leverages a deep network to extract video features and explicitly
estimates an ordered temporal alignment path to match the frames of two videos.
HyRSM~\cite{HyRSM} proposes to explore task-specific semantic correlations across videos and designs a bidirectional Mean Hausdorff Metric (Bi-MHM) to align frames.
%
Though these works have obtained significant results, there are still two limitations:
first, existing standard metric-based techniques mainly focus on local frame-level alignment and are considered limited since the essential global information is not explicitly involved.
As shown in Figure~\ref{fig:Motivation}, local frame-level  metrics can be easily affected by co-existing similar video frames.
We argue that it would be beneficial to achieve accurate matching if the local frame features can predict the global context in few-shot classification;
%
%
%
%
%
second, motion dynamics are widely regarded as a vital role in the field of video understanding~\cite{rui2000segmenting,niebles2010modeling,jain2013better,STM,wang2022context,cen2023devil}, while the existing few-shot methods do not explicitly explore the rich motion cues between frames for the matching procedure, resulting in a sub-optimal performance.
In the literature~\cite{TSN,Kinetics}, traditional action recognition works introduce motion information by feeding optical flow or frame difference into an additional deep network, which leads to non-negligible computational overhead.
Therefore, an efficient motion compensation method should be introduced to achieve comprehensive few-shot matching.
%
%
%
%

Inspired by the above observations, we develop a motion-augmented long-short contrastive learning (MoLo) method to jointly model the global contextual information and motion dynamics.
More specifically, to explicitly integrate the global context into the local matching process, 
we apply a long-short contrastive objective to enforce frame features to predict the global context of the videos that belong to the same class. 
For motion compensation, we design a motion autodecoder to explicitly extract motion features between frame representations by reconstructing pixel motions, \eg, frame differences.
%
In this way, our proposed MoLo enables efficient and comprehensive exploitation of temporal contextual dependencies and motion cues for accurate few-shot action recognition.
%
Experimental results on multiple widely-used benchmarks demonstrate that our MoLo outperforms other advanced few-shot techniques and achieves state-of-the-art performance.

In summary, our contributions can be summarized as follows: 
(1) We propose a novel MoLo method for few-shot action recognition, aiming to
better leverage the global context and motion dynamics.
(2) We further design a long-short contrastive objective to reinforce local frame features to perceive comprehensive global information and a motion autodecoder to 
explicitly extract motion cues.
(3) We conduct extensive experiments across five widely-used benchmarks to validate the effectiveness of the proposed MoLo.  
The results demonstrate that MoLo significantly outperforms baselines and achieves state-of-the-art performance.

%

\section{Related Work}
\label{sec:related}

%
%
In this section, we will briefly review the works closely related to this paper, including few-shot image classification, motion learning, and few-shot action recognition. 

\vspace{+1mm}
\noindent \textbf{Few-shot image classification. }
Identifying unseen classes using only a few labels, known as few-shot learning~\cite{few-shot_feifei}, is an important research direction in computer vision.
Existing few-shot learning techniques can generally be divided into three groups: 
augmentation-based, gradient optimization, and metric-based methods.
Augmentation-based approaches usually learn to generate samples to alleviate the data scarcity dilemma, mainly including hallucinating additional training examples~\cite{data-aug-1,data-aug-2,data-aug-4,data-aug-5} and adversarial generation~\cite{data-aug-3,data-aug-6}.
Gradient optimization methods~\cite{MAML,MAML-1,MAML-2,MAML-3,MAML-4,MAML-5} attempt to modify the network optimization process so that the model can be quickly fine-tuned to near the optimal point.
For example, MAML~\cite{MAML} explicitly trains the parameters of the network to produce good generalization performance on the current task with a small
number of gradient steps.
Metric-based paradigm~\cite{prototypical} is widely adopted, leveraging a visual encoder to map images into an embedding space and learning a similarity function.
This type of methods generally performs few-shot matching to classify query samples through Euclidean distance~\cite{prototypical,oreshkin2018tadam,wang2019simpleshot,yoon2019tapnet}, cosine distance~\cite{MatchNet,qiao2018few}, and learnable metrics~\cite{RelationNet,hao2019collect,wang2020cooperative}.
Our algorithm also falls into the metric-based line while focusing on taking advantage of temporal context and motion information for accurate few-shot action recognition.

\vspace{+1mm}
\noindent \textbf{Motion Learning.}
How to efficiently learn motion cues in videos is a widely researched problem in the community. 
Previous methods typically exploit optical flow~\cite{wang2013action,piergiovanni2019representation,simonyan2014two,wang2022context,zhu2006action,wang2016robust,wang2021weakly} or frame difference~\cite{lee2011vision,zhang2012slow,TDN} to explicitly inject video dynamic cues for action recognition.
Due to the high acquisition cost of optical flow~\cite{beauchemin1995computation,horn1981determining,lee2018motion} and the need for additional networks to extract the motion features~\cite{TSN,Kinetics}, recent approaches begin to study how to extract motion features with only raw frame input and achieve promising results~\cite{ji20123d,tran2018closer,fan2020rubiksnet,STM,wang2022long,dave2022tclr,wang2020video,kwon2020motionsqueeze,kwon2021learning}.
In this paper, we also focus on using raw frames to extract motion information. 
Different from previous methods, the motion feature in our MoLo is used to facilitate video alignment, and we design a motion autodecoder that learns to recover frame differences to explicitly extract motion dynamics in a unified network without the need for multiple separate networks.
\begin{figure*}[t]
  \centering
   \includegraphics[width=0.96\linewidth]{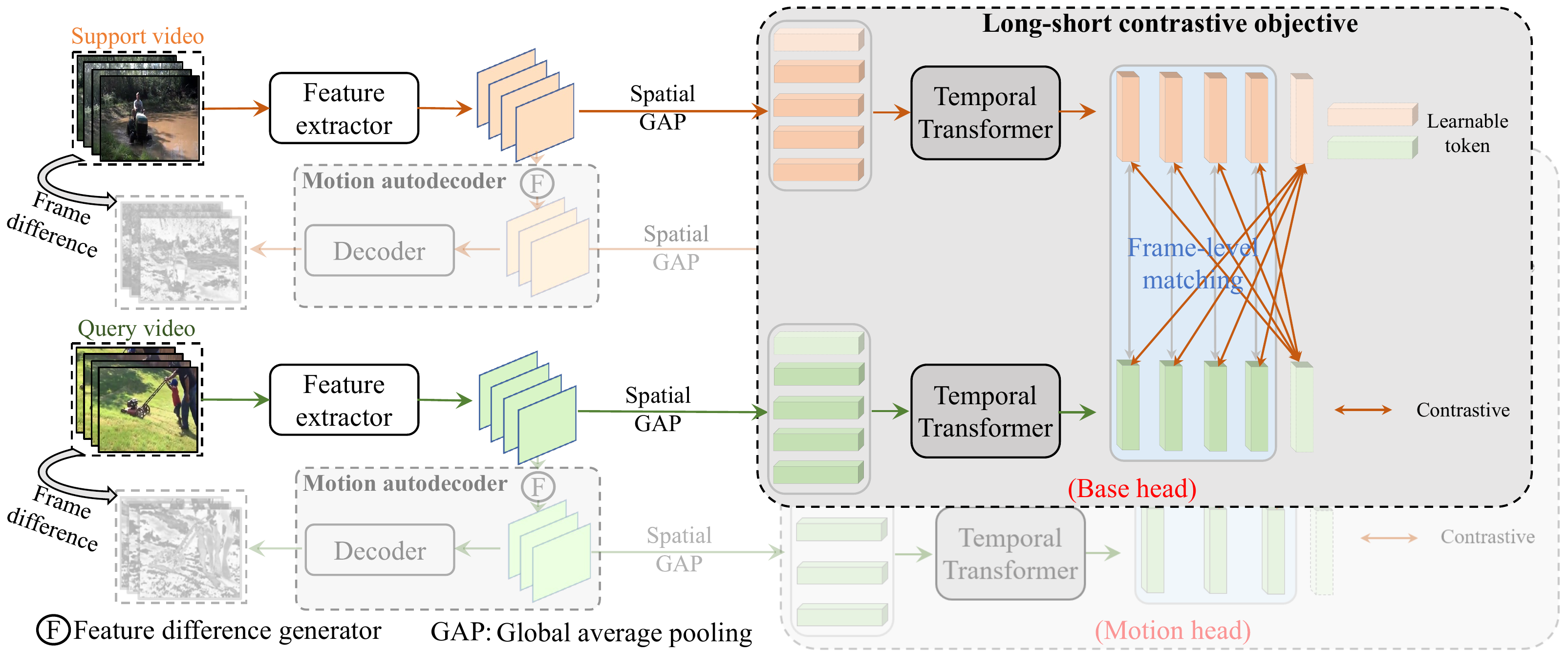}
\vspace{-4mm}
   \caption{Overview of the proposed MoLo.
   Given the support and query videos, we first leverage a feature extractor to encode frame features. 
 Subsequently, we feed these frame features into the base head to perform support-query frame-level metric and apply the long-short contrastive objective to improve temporal context awareness during the matching process.
   Besides, a motion autodecoder is used to recover pixel motions, \ie, frame differences, and the obtained motion features are then entered into a motion head similar to the base head for few-shot matching.
   Finally, the query video is classified by fusing the matching results of the two heads.
   For the convenience of illustration, other support videos in a few-shot task are not displayed in the figure.
%
   %
   }
   \label{fig:network}
   \vspace{-4mm}
\end{figure*}

\vspace{+1mm}
\noindent \textbf{Few-shot action recognition.}
Compared to images, videos contain richer spatio-temporal context interactions and temporal variations~\cite{poppe2010survey,wang2021self,OSS-metric,wang2021proposal}. 
Existing few-shot action recognition methods~\cite{OTAM,wang2023clip,wang2023hyrsm++} mainly belong to the metric-based meta-learning paradigm~\cite{prototypical} to temporally match frames, due to its simplicity and effectiveness.
Among them,
CMN~\cite{CMN,CMN-J} proposes a memory structure and a multi-saliency algorithm to encode variable-length video sequences
into fixed-size matrix representations for video matching. 
OTAM~\cite{OTAM} designs a differentiable dynamic time warping~\cite{DTW} algorithm to temporally
align two video sequences.
ARN~\cite{ARN-ECCV} builds on a 3D backbone~\cite{C3D} to capture long-range temporal dependencies and learn permutation-invariant representations.
TRX~\cite{TRX} adopts an attention mechanism to align each
query sub-sequence against all sub-sequences in the support set and then fuses their matching scores for classification. 
ITANet~\cite{ITANet} and STRM~\cite{STRM} introduce joint spatio-temporal modeling techniques for robust few-shot matching.
HyRSM~\cite{HyRSM} and MTFAN~\cite{MTFAN} emphasize the importance of learning task-specific features.
Both Nguyen~\etal~\cite{nguyen2022inductive} and Huang~\cite{huang2022compound} employ multiple similarity functions to compare videos accurately.
Despite the remarkable performance, most of these methods are generally limited to local matching without explicitly considering temporal contextual information and motion cues.

\section{Method}
\label{sec:method}

In this section, we first introduce the problem setup of the few-shot action recognition task.
%
%
Then, we will elaborate on the proposed MoLo method.

\subsection{Problem definition}

Given a training set $\mathcal{D}_{train}=\{ (v_{i},y_{i}), y_{i} \in \mathcal{C}_{train} \} $ and a test set $\mathcal{D}_{test}=\{ (v_{i},y_{i}), y_{i} \in \mathcal{C}_{test} \} $, we usually train a deep network on the training set and then verify the performance on the test set.
Traditional supervised learning tasks assume that training and testing samples share the same set of classes, \ie, $\mathcal{C}_{train} = \mathcal{C}_{test}$.
In contrast, the objective of few-shot action recognition is to identify unseen classes using only a few samples during testing, \ie, $\mathcal{C}_{train}\cap\mathcal{C}_{test}=\emptyset$, and verify the generalization of the trained model.
Following previous approaches~\cite{OTAM,TRX,HyRSM}, we adopt the episode-based meta-learning strategy~\cite{MatchNet} to optimize the network.
In each episode, there is a support set $S$ containing $N$ classes and $K$ videos per class (called the $N$-way $K$-shot task) and a query set $Q$ containing query samples to be classified.
Few-shot action recognition models are expected
to learn to predict labels for videos in the
query set with the guidance of the support set.
For inference, a large number of episodic tasks will be randomly sampled on the test set, and the average accuracy is used to report the few-shot performance of the learned model.
%

%

\subsection{MoLo}

\textbf{Overall architecture.}
An overview of our framework is provided in Figure~\ref{fig:network}.
For clear description, we take the $N$-way 1-shot task as an example to illustrate our approach.
%
In an $N$-way 1-shot episodic task, the support set $S=\{s_{1}, s_{2},...,s_{N} \}$ contains $N$ video samples, where $s_{i}\in \mathbb{R} ^{T\times 3 \times H\times W}$  and $T$ is the number of sparsely sampled video frames to obtain the video representation.
Given a query video $q\in \mathbb{R} ^{T\times 3 \times H\times W}$, the goal is to accurately classify $q$ as a category in the support set.
Like previous methods~\cite{OTAM,HyRSM}, we first adopt a deep feature extraction network~\cite{Resnet} to encode the video into feature sequence. Thus we can obtain the support features $F_{S}=\{f_{s_1},f_{s_2},...,f_{s_N} \}$ and query feature $f_{q}$, where $f_{i}=\{f^{1}_{i},f^{2}_{i},...,f^{T}_{i}  \}, f^{j}_{i}\in \mathbb{R} ^{C \times H_{f}\times W_{f}}$ and $C$ is the channel number.
Subsequently,  the obtained features are fed into a base head with the long-short contrastive objective for frame matching. 
Meanwhile, in order to explicitly extract motion information for few-shot matching, we also
design a motion autodecoder to reconstruct the frame differences and then input the resulting motion features into a motion head similar to the base head.
%
Based on the output matching scores from the above two heads, we can merge them for query classification.

\begin{figure}[t]
  \centering
   \includegraphics[width=0.99\linewidth]{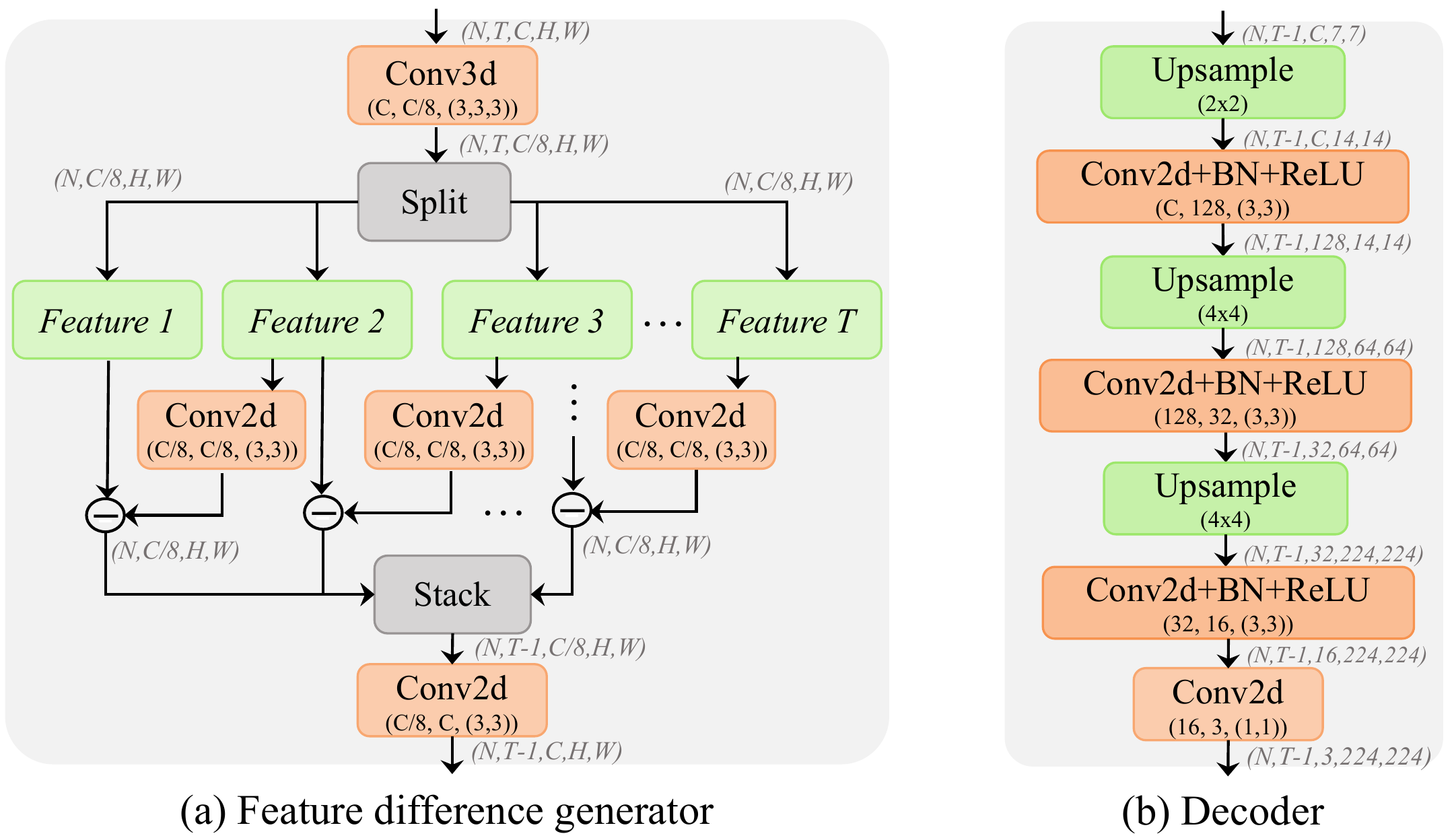}
\vspace{-3mm}
   \caption{The architecture details of the proposed motion autodecoder, including a feature difference generator and a decoder.
   (a) The feature difference generator computes the feature difference between adjacent video frames after a 3D convolution to extract motion information.
   (b) The decoder involves stacking several upsampling operations to reconstruct the frame differences.
   }
   \label{fig:motion_module}
   \vspace{-3mm}
\end{figure}

\textbf{Long-short contrastive objective.}
Given the features $F_{S}$ and $f_{q}$, a spatial global-average pooling is applied to collapse the spatial dimension.
After that, we append a learnable token to the video feature sequence and send the features to a temporal Transformer for temporal information aggregation.
We denote the output features as $\tilde{f_{i}}=\{\tilde{f}^{token}_{i},\tilde{f}^{1}_{i},...,\tilde{f}^{T}_{i}  \}$.
The process can be formulated as:
\begin{equation}\label{eq: 1}
    \tilde{f_{i}} = \mathrm{Tformer}([{f}^{token}, \mathrm{GAP}(f^{1}_{i},...,f^{T}_{i}) ] + f_{pos})
\end{equation}
where $\mathrm{Tformer}$ is the temporal Transformer, ${f}^{token}\in \mathbb{R} ^{ C}$ is a learnable token shared cross videos, and $f_{pos}\in \mathbb{R} ^{(T+1)\times C}$ means the position embeddings to encode the relative position relationship.
Note that the learnable token is used to aggregate global information adaptively, and previous methods~\cite{ViT,vivit,wang2021oadtr} have also shown the effectiveness of this manner.
In this way, the output feature $\tilde{f}^{token}_{i}$ corresponding to the learnable token has the global representation ability.
We then show how we leverage the global property of the token feature for robust video matching.

Most existing methods~\cite{OTAM,ITANet,HyRSM} usually classify query video sample $\tilde{f_{q}}$ based on the frame-level alignment metric, which can be expressed as:
\begin{equation}\label{eq: 2}
    D(\tilde{f_{i}}, \tilde{f_{q}}) = \mathcal{M} ([\tilde{f}^{1}_{i},...,\tilde{f}^{T}_{i} ], [\tilde{f}^{1}_{q},...,\tilde{f}^{T}_{q}])
\end{equation}
where $\mathcal{M}$ is a frame-level metric, such as OTAM~\cite{OTAM} and Bi-MHM~\cite{HyRSM}.
However, the above metric merely considers the local frame level, ignoring explicitly enhancing global perceptual capabilities of frame features in the matching process.
Intuitively, if two videos belong to the same class, the distance between the global feature of one video and the frame feature of the other video is closer than other videos of different categories.
We argue that the global context is much critical for local frame-level matching, especially in the presence of similar co-existing frames.
To this end, we leverage a long-short contrastive loss based on ~\cite{MIL-NCE} to improve the global context awareness of frame features:
\begin{align}
\nonumber
&\mathcal{L}^{base}_{LG} = -\mathrm{log} \frac{ { \sum\limits_{ i}} \mathrm{sim}(\tilde{f}^{token}_{q}, \tilde{f}^{i}_{p})}{{ \sum\limits_{ i}} \mathrm{sim}(\tilde{f}^{token}_{q}, \tilde{f}^{i}_{p})+{ \sum\limits_{ j\ne p}}\sum\limits_{ i}\mathrm{sim}(\tilde{f}^{token}_{q}, \tilde{f}^{i}_{j}) } 
\\ 
&-\mathrm{log} \frac{ { \sum\limits_{ i}} \mathrm{sim}(\tilde{f}^{i}_{q}, \tilde{f}^{token}_{p})}{{ \sum\limits_{ i}} \mathrm{sim}(\tilde{f}^{i}_{q}, \tilde{f}^{token}_{p})+{ \sum\limits_{ j\ne q}}\sum\limits_{ i}\mathrm{sim}(\tilde{f}^{i}_{j}, \tilde{f}^{token}_{p}) } 
\end{align}
where $\tilde{f}_{q}$ is the query feature, $\tilde{f}_{p}$ is video representations of the same class as the query in an episodic task, and $\tilde{f}_{j}$ is the negative sample that belongs to the other category.
The loss function above is expected to maximize the sum of all positive paired similarity scores and minimize the sum of all negative paired similarity scores.
%
%
%
In this way, we explicitly force local features to perceive the global temporal context, enabling more robust few-shot frame alignment.
%


\begin{table*}[t]
\centering
\small
\tablestyle{6pt}{0.93}
\setlength{
    \tabcolsep}{
    2.3mm}{
\begin{tabular}
{l|c|ccccc|ccccc}
\hline
			
\hspace{-0.5mm}  \multirow{2}{*}{Method} \hspace{2mm} & \multicolumn{1}{c|}{\multirow{2}{*}{Reference}} & \multicolumn{5}{c|}{{SSv2-Full}}  & \multicolumn{5}{c}{{Kinetics}}  \\
& \multicolumn{1}{c|}{} 
& \multicolumn{1}{l}{1-shot} & 2-shot  & 3-shot  & 4-shot   & 5-shot 
&  \multicolumn{1}{l}{1-shot} & 2-shot  & 3-shot  & 4-shot   & 5-shot  \\ \shline

MatchingNet~\cite{MatchNet}    & NeurIPS'16
& -  & -  & -   & -   & -   
& 53.3   & 64.3  & 69.2  & 71.8  & 74.6  \\
MAML~\cite{MAML}    & ICML'17
& -  & -  & -   & -   & -   
& 54.2   & 65.5  & 70.0  & 72.1  & 75.3  \\
Plain CMN~\cite{CMN}    & ECCV'18
& -  & -  & -   & -   & -   
& 57.3   & 67.5  & 72.5  & 74.7  & 76.0  \\
CMN++~\cite{CMN}    & ECCV'18
& 34.4  & -  & -   & -   & 43.8   
& -   & -  & -  & -  & -  \\
TRN++~\cite{TRN-ECCV}    & ECCV'18
& 38.6  & -  & -   & -   & 48.9   
& -   & -  & -  & -  & -  \\
TARN~\cite{TARN}    & BMVC'19
& -  & -  & -   & -   & -   
& 64.8   & -  & -  & -  & 78.5  \\
CMN-J~\cite{CMN-J}    & TPAMI'20
& -  & -  & -   & -   & -   
& 60.5   & 70.0  & 75.6  & 77.3  & 78.9  \\
ARN~\cite{ARN-ECCV}    & ECCV'20
& -  & -  & -   & -   & -   
& 63.7   & -  & -  & -  & 82.4  \\
OTAM~\cite{OTAM}    & CVPR'20
& 42.8  & 49.1  & 51.5   & 52.0   & 52.3   
& \hspace{0.6mm} 72.2$^{*}$   & 75.9  & 78.7  & 81.9  &  \hspace{0.6mm} 84.2$^{*}$  \\
ITANet~\cite{ITANet}    & IJCAI'21
& 49.2  & 55.5  & 59.1   & 61.0   & 62.3   
& 73.6   & -  & -  & -  & 84.3  \\
TRX ($\Omega{=}\{1\}$)~\cite{TRX}    & CVPR'21     & 38.8     & 49.7   & 54.4     & 58.0      & 60.6   & 63.6  & 75.4  & 80.1  & 82.4  & 85.2   \\ 
TRX ($\Omega{=}\{2,3\}$)~\cite{TRX}  & CVPR'21          &  42.0   &  53.1   &  57.6   &   {61.1}   &   {64.6}   & 63.6  & {76.2}  & {81.8}  & {83.4}  & {85.9}  \\  
TA$^{2}$N~\cite{TA2N}    & AAAI'22
& 47.6  & -  & -   & -   & 61.0   
& 72.8   & -  & -  & -  & 85.8  \\
MTFAN~\cite{MTFAN}    & CVPR'22
& 45.7  & -  & -   & -   & 60.4   
& \textbf{74.6}   & -  & -  & -  & \textbf{87.4}  \\
STRM~\cite{STRM}    & CVPR'22
& 43.1 & 53.3 & 59.1 & 61.7 & 68.1
& 62.9 & 76.4 & 81.1 & 83.8 & 86.7 \\
HyRSM~\cite{HyRSM}    & CVPR'22
& 54.3  & \underline{62.2}  & \underline{65.1}   & \underline{67.9}   & 69.0   
& 73.7   & 80.0  & \underline{83.5}  & \underline{84.6}  & 86.1  \\
Bi-MHM~\cite{HyRSM}    & CVPR'22
& \hspace{1.2mm}44.6$^{*}$  & \hspace{1.2mm}49.2$^{*}$ & \hspace{1.2mm}53.1$^{*}$   & \hspace{1.2mm}54.8$^{*}$   & \hspace{1.2mm}56.0$^{*}$   
& \hspace{1.2mm}72.3$^{*}$   & \hspace{1.2mm}77.2$^{*}$  & \hspace{1.2mm}81.1$^{*}$  & \hspace{1.2mm}84.1$^{*}$  &\hspace{1.2mm} 84.5$^{*}$  \\
Nguyen~\etal~\cite{nguyen2022inductive}    & ECCV'22
& 43.8  & -  & -   & -   & 61.1   
& \underline{74.3}   & -  & -  & -  & \textbf{87.4}  \\
Huang~\etal~\cite{huang2022compound}    & ECCV'22
& 49.3  & -  & -   & -   & 66.7   
& 73.3   & -  & -  & -  & 86.4 \\
HCL~\cite{HCL}    & ECCV'22
& 47.3  & 54.5  & 59.0   & 62.4   & 64.9   
& 73.7   & 79.1  & 82.4  & 84.0  & 85.8 
\\ \shline
\rowcolor{Gray}
\textbf{MoLo} (OTAM) & -
& \underline{55.0}  & 61.8  & 64.8   & 67.7   & \underline{69.6}   
& 73.8   & \underline{80.2}  & 83.1  & 84.2  & 85.1  \\ 
\rowcolor{Gray}
\textbf{MoLo} (Bi-MHM) & -
& \textbf{56.6}  & \textbf{62.3}  & \textbf{67.0}   & \textbf{68.5}   & \textbf{70.6}  
& 74.0   & \textbf{80.4}  & \textbf{83.7}  & \textbf{84.7}  & 85.6  \\ 
\hline
\end{tabular}
}
\vspace{-3mm}
\caption{Comparison with recent state-of-the-art few-shot action recognition methods on the SSv2-Full and Kinetics datasets under the 5-way setting. 
The experimental results are reported as the shot increases from 1 to 5.
"-" indicates the result is not available in published works.
The best results are bolded and the underline means the second best performance.
``${*}$"  stands for the results of our implementation.
}
\label{tab:compare_SOTA_1}
\vspace{-4mm}
\end{table*}

\textbf{Motion autodecoder.}
To further incorporate motion information into the few-shot matching process, we design a motion autodecoder consisting of a feature difference generator and a decoder.
The motion autodecoder explicitly extract motion features by operating between frame features.
%
%
Specifically, we modify the differential operator in ~\cite{STM} to generate features difference, as shown in Figure~\ref{fig:motion_module}(a).
The formula is expressed as follows:
\begin{equation}\label{eq: motion1}
    f^{{\prime} 1}_{i},...,f^{{\prime} T-1}_{i} = \mathcal{F} (f^{1}_{i},...,f^{T}_{i})
\end{equation}
where $\mathcal{F}$ means the feature difference generator, and $f^{{\prime} j}_{i}$ is the output motion feature.
%
%
Through the above feature difference generator, we intuitively obtain the motion features between frames.
To further explicitly supervise motion feature generation, inspired by MAE~\cite{MAE}, we design a decoder (Figure~\ref{fig:motion_module}(b)) to reconstruct pixel motion dynamics.
Subsequently, we input the obtained motion features into a motion head with long-short contrastive objective for video matching.
Like the base head, in motion head we first adopt a spatial global-average pooling to motion features and concatenate a learnable token, which are then fed into a temporal Transformer for temporal modeling. We denote the resulting features as $\tilde{f}^{{\prime}}_{i}= \{\tilde{f}^{{\prime} token}_{i},\tilde{f}^{{\prime} 1}_{i},...,\tilde{f}^{ {\prime} T-1}_{i}\}$.
For local matching, we apply a frame-level metric to these features:
\begin{equation}\label{eq: motion2}
    D(\tilde{f^{{\prime}}_{i}}, \tilde{f^{{\prime}}_{q}}) = \mathcal{M} ([\tilde{f}^{{\prime}1}_{i},...,\tilde{f}^{{\prime}T-1}_{i} ], [\tilde{f}^{{\prime}1}_{q},...,\tilde{f}^{{\prime}T-1}_{q}])
\end{equation}
where $D(\tilde{f^{{\prime}}_{i}}, \tilde{f^{{\prime}}_{q}})$ is the distance between the support motion feature $\tilde{f^{{\prime}}_{i}}$ and query motion feature $\tilde{f^{{\prime}}_{q}}$.
%
%
Similar to $\mathcal{L}^{base}_{LG}$, we also apply a loss $\mathcal{L}^{motion}_{LG}$ to the motion features.

%
We express the distance between support and query videos as the weighted sum of the distances obtained by the base head and motion head:
\begin{equation}\label{eq: distance}
D_{i,q} = D(\tilde{f_{i}}, \tilde{f_{q}}) +\alpha D(\tilde{f^{{\prime}}_{i}}, \tilde{f^{{\prime}}_{q}})
\end{equation}
where $\alpha$ is a balance coefficient.
Then, we can use the output support-query distances as logits for classification.
%
%
During the training process, the overall framework is trained end-to-end, and the final loss can be denoted as:
\begin{equation}\label{eq: loss}
\mathcal{L} = \mathcal{L}_{CE}+{\lambda}_1     (\mathcal{L}^{base}_{LG} +     \mathcal{L}^{motion}_{LG}) + {\lambda}_2 \mathcal{L}_{Recons}
\end{equation}
where $\mathcal{L}_{CE}$ is the cross-entropy loss over the support-query distances based on the ground-truth label, ${\lambda}_1$ and ${\lambda}_2$ are balanced factors, and $\mathcal{L}_{Recons}$ represents the L2 loss of reconstructing frame differences.
For few-shot inference, we can leverage the distance values of support-query video pairs in Equation~\ref{eq: distance} to classify query samples based on the nearest neighbor rule~\cite{prototypical}, and the decoder can be discarded.


%
\begin{table*}[t]
\centering
\small
\tablestyle{6pt}{1.0}
\setlength{
    \tabcolsep}{
    3.0mm}{
\begin{tabular}
{l|c|ccc|ccc|ccc}
\hline
			
\hspace{-0.5mm}  \multirow{2}{*}{Method} \hspace{2mm} & \multicolumn{1}{c|}{\multirow{2}{*}{Reference}} & \multicolumn{3}{c|}{{UCF101}}  &\multicolumn{3}{c|}{{SSv2-Small}}  & \multicolumn{3}{c}{{HMDB51}}  \\
& \multicolumn{1}{c|}{} 
& \multicolumn{1}{l}{1-shot} & 3-shot  & 5-shot & \multicolumn{1}{l}{1-shot} & 3-shot   & 5-shot 
& \multicolumn{1}{l}{1-shot}& 3-shot    & 5-shot  \\ \shline
MatchingNet~\cite{MatchNet}    & NeurIPS'16
& -  & -  & -   
& 31.3   & 39.8   & 45.5   
& -  & -  & -    \\
MAML~\cite{MAML}    & ICML'17
& -  & -  & -   
& 30.9   & 38.6   & 41.9   
& -  & -  & -    \\
Plain CMN~\cite{CMN}    & ECCV'18
& -  & -  & -   
& 33.4   & 42.5   & 46.5   
& -  & -  & -    \\
CMN-J~\cite{CMN-J}    & TPAMI'20
& -  & -  & -   
& 36.2   & 44.6   & 48.8   
& -  & -  & -    \\
ARN~\cite{ARN-ECCV}    & ECCV'20
& 66.3  & -  & 83.1   
& -   & -   & -   
& 45.5  & -  & 60.6    \\
OTAM~\cite{OTAM}    & CVPR'20
& 79.9  & 87.0  & 88.9   
& 36.4  & 45.9  & 48.0   
& 54.5  & 65.7  & 68.0    \\
ITANet~\cite{ITANet}    & IJCAI'21
& -  & -  & -   
& 39.8  & 49.4  & 53.7  
& -  & -  & -    \\
TRX~\cite{TRX}    & CVPR'21
& 78.2  & 92.4  & \underline{96.1}   
& 36.0  & 51.9  & \hspace{1.2mm} \underline{56.7}$^{*}$  
& 53.1  & 66.8  & 75.6    \\
TA$^{2}$N~\cite{TA2N}    & AAAI'22
& 81.9  & -  & 95.1   
& -  & -  & -   
& 59.7  & -  & 73.9    \\
MTFAN~\cite{MTFAN}    & CVPR'22
& 84.8  & -  & 95.1   
& -  & -  & -  
& 59.0  & -  & 74.6    \\
STRM~\cite{STRM}    & CVPR'22
& 80.5 & 92.7 & \textbf{96.9}
& 37.1 & 49.2 & 55.3
& 52.3 & 67.4 & \underline{77.3}   \\
HyRSM~\cite{HyRSM}    &  CVPR'22
& 83.9  & 93.0  & 94.7   
& 40.6  & \underline{52.3}  & 56.1  
& \underline{60.3}  & \underline{71.7}  & 76.0    \\
Bi-MHM~\cite{HyRSM}    & CVPR'22
& \hspace{1.2mm}81.7$^{*}$  & \hspace{1.2mm}88.2$^{*}$ & \hspace{1.2mm}89.3$^{*}$   & \hspace{1.2mm}38.0$^{*}$   & \hspace{1.2mm}47.6$^{*}$   
& \hspace{1.2mm}48.9$^{*}$   & \hspace{1.2mm}58.3$^{*}$  & \hspace{1.2mm}67.1$^{*}$  & \hspace{1.2mm}69.0$^{*}$ \\
Nguyen~\etal~\cite{nguyen2022inductive}    & ECCV'22
& 84.9  & -  & 95.9   
& -  & -  & -  
& 59.6  & -  & {76.9}    \\
Huang~\etal~\cite{huang2022compound}    & ECCV'22
& 71.4  & -  & 91.0   
& 38.9  & -  & \textbf{61.6}  
& 60.1  & -  & {77.0}    \\ 
HCL~\cite{HCL}    & ECCV'22
& 82.5  & 91.0  & 93.9   
& 38.7  & 49.1  & 55.4  
& 59.1  & 71.2  & 76.3    \\ 

\shline
\rowcolor{Gray}
\textbf{MoLo} (OTAM)    & -
& \underline{85.4}  & \underline{93.4}  & 95.1   
& \underline{41.9}  & 50.9  & 56.2  
& 59.8  & 71.1  & 76.1    \\
\rowcolor{Gray}
\textbf{MoLo} (Bi-MHM)    & -
& \textbf{86.0}  & \textbf{93.5}  & 95.5   
& \textbf{42.7}  & \textbf{52.9}  & {56.4}  
& \textbf{60.8}  & \textbf{72.0}  & \textbf{77.4}    \\
\hline
\end{tabular}
}
\vspace{-2mm}
\caption{Comparison with state-of-the-art few-shot action recognition methods on UCF101, SSv2-Small, and HMDB51 in terms of  1-shot, 3-shot, and 5-shot classification accuracy.
"-" stands for the result is not available in published works.
The best results are bolded in black, and the underline represents the second best result.
``${*}$"  indicates the results of our implementation.
}
\label{tab:compare_SOTA_2}
\vspace{-2mm}
\end{table*}

\section{Experiments}
\label{sec:experiments}
In this section, we first introduce the experimental settings of our MoLo in Section~\ref{sec:datasets} and then compare our approach with previous state-of-the-art methods on multiple commonly used benchmarks in Section~\ref{sec:comparison_with_SOTA}. Finally, comprehensive ablation studies of MoLo are provided  in Section~\ref{sec:ablation_study} to demonstrate the effectiveness of each module.

\subsection{Datasets and experimental setup}
\label{sec:datasets}
\noindent \textbf{Datasets.} Our experiments are conducted on five commonly used few-shot datasets, including SSv2-Full~\cite{SSV2}, SSv2-Small~\cite{SSV2}, Kinetics~\cite{Kinetics}, UCF101~\cite{UCF101}, and HMDB51~\cite{HMDB51}.
%
%
For SSv2-full and SSv2-Small, we follow the spilt settings from ~\cite{OTAM} and ~\cite{CMN}, which randomly choose 64 classes from the original dataset~\cite{SSV2} as the training set and 24 classes as the test set.
The difference between SSv2-Full and SSv2-Small is that the former contains all samples of each category, while the latter selects 100 samples per category.
Kinetics used in the few-shot setup~\cite{OTAM,HyRSM} is also a subset of the original dataset~\cite{Kinetics}. 
The UCF101~\cite{UCF101} dataset contains 101 action classes, and we adopt the few-shot split as ~\cite{ARN-ECCV,MTFAN,HyRSM}.
For HMDB51~\cite{HMDB51}, the common practice is to split 31 classes for training and 10 classes for testing.

\noindent \textbf{Implementation details.}
For a fair comparison with existing methods~\cite{CMN,OTAM,TRX}, our approach employs a ResNet-50~\cite{Resnet} pre-trained on ImageNet~\cite{imagenet} as the basic feature extractor and removes the last global-average pooling layer to retain spatial motion information. 
We adopt Adam~\cite{adam} to optimize our MoLo end-to-end and apply an auxiliary semantic loss to stabilize the training process as in ~\cite{ITANet,few-shot-regular,HyRSM,few-shot-regular-2}.
Like previous methods~\cite{OTAM,TRX}, 8 frames are uniformly sparsely sampled, i.e. $T=8$, to obtain the representation of the whole video.
We also implement standard data augmentation such as random crop and color jitter in the training stage, and a $224 \times 224$ region is cropped at the center of each frame for few-shot testing.
For many-shot inference (\eg, 5-shot), we simply embrace the averaged prototype paradigm~\cite{prototypical} to classify query video samples.
To evaluate few-shot performance on each benchmark,
we randomly construct 10,000 episodes from the test set and report the average classification accuracy.

\subsection{Comparison with state-of-the-art methods}
\label{sec:comparison_with_SOTA}
We compare the performance of MoLo with existing state-of-the-art methods
under the 5-way $K$-shot setting, with $K$ varying from 1 to 5 on SSV2-Full and Kinetics.
As displayed in Table~\ref{tab:compare_SOTA_1}, 
two commonly used alignment metrics are adopted to verify the effectiveness of our framework, \ie, OTAM~\cite{OTAM} and Bi-MHM~\cite{HyRSM}.
%
From the results, we can observe the following:
(1) Based on Bi-MHM, our MoLo consistently outperforms all prior works on the SSv2-Full dataset under various shot settings and achieves comparable results on Kinetics, which validates our motivation to introduce global information and motion cues for reliable video matching.
%
(2) Compared with the vanilla OTAM, our method can significantly improve performance. For instance, MoLo improves the performance from 42.8\% to 55.0\% under the 1-shot SSv2-full setting and obtains 1.6\% 1-shot performance gain on Kinetics. The same phenomenon holds for Bi-MHM, which demonstrates the robustness of our framework to the type of local frame metrics.
%
(3) The gains of our MoLo on the SSv2-Full dataset are generally more significant than that on the Kinetics dataset. We attribute this to the fact that the SSv2-Full dataset involves a large amount of temporal reasoning, and explicitly introducing global temporal context and motion compensation yields substantial benefits on this dataset, while the Kinetics dataset is more biased towards appearance scenes.

\begin{table}[t]
\centering
\small
\tablestyle{6pt}{1.0}
\setlength{
    \tabcolsep}{
    1.0mm}{
\begin{tabular}
{cc|cc|cc}
\hline
			
  \multicolumn{2}{c|}{}  & \multicolumn{2}{c|}{Head}  & 
\multicolumn{2}{c}{{SSv2-Full}}  
\\

\multicolumn{1}{l}{Long-short contrastive}  & Autodecoder & \multicolumn{1}{l}{Base} & Motion  & \multicolumn{1}{l}{1-shot} & 5-shot  
\\ \shline

%
 &  &\checkmark & 
& 44.6 & 56.0     \\ 
 &  & & \checkmark
& 46.3 & 60.6     \\ 
\shline
%
%
\checkmark &  &\checkmark & 
& 52.2 & 68.0     \\ 
\checkmark & \checkmark &\checkmark & 
& 53.2 & 68.1    \\ 
 & \checkmark  &  & \checkmark 
& 47.8 & 61.8     \\ 
\checkmark & \checkmark  &  & \checkmark 
& 53.9 & 69.7     \\ 
\shline
  &  & \checkmark& \checkmark
& 49.2 & 63.4    \\ 
\checkmark &  & \checkmark &  \checkmark
& 53.3 & 68.2    \\ 
 & \checkmark  & \checkmark &  \checkmark
& 53.2 & 68.1    \\ 
\checkmark & \checkmark & \checkmark& \checkmark
& \textbf{56.6} & \textbf{70.6}    \\ 
\hline
\end{tabular}
}
\vspace{-1mm}
\caption{Ablation study on SSv2-Full under 5-way 1-shot
and 5-way 5-shot settings.
The top line represents the baseline Bi-MHM.
To avoid confusion, note that the ``motion head without autodecoder" setting contains the feature difference generator by default.
%
}
\label{tab:ablation}
\vspace{-2mm}
\end{table}
%
In order to further verify our MoLo, we also compare with state-of-the-art methods on the UCF101, HMDB51, and SSv2-Small datasets, and the results are shown in Table~\ref{tab:compare_SOTA_2}.
The conclusions are basically consistent with those in Table~\ref{tab:compare_SOTA_1}. Notably, our method outperforms other methods in most cases (\eg, 1-shot and 3-shot). However, the performance on 5-shot lags slightly behind the recent advanced methods on UCF101 and SSv2-Small, since our MoLo is a general framework for different shot settings and is not specially designed for 5-shot like TRX~\cite{TRX} and STRM~\cite{STRM}.

\subsection{Ablation study}
\label{sec:ablation_study}
For the convenience of comparison, we choose Bi-MHM~\cite{HyRSM} as our baseline and add each component into Bi-MHM to verify the rationality of the module design.

\vspace{1mm}
\noindent \textbf{Analysis of network components.}
We present a detailed ablation analysis of model elements, and the experimental results are summarized in Table~\ref{tab:ablation}.
Based
on the results, we can observe that the long-short contrastive objective and motion autodecoder play a key role in boosting performance.
In particular, 1-shot performance can be improved from 44.6\% to 52.2\% by adding the long-short contrastive objective to baseline BiMHM, reflecting the importance of incorporating global context into the local matching process.
The performance gain is also significant by introducing the motion autodecoder.
In addition, the experimental results show the complementarity among different modules, indicating the rationality of our model structure.
By integrating all modules, our method achieves 56.6\% and 70.6\% performance under 1-shot and 5-shot settings, respectively.
In Table~\ref{tab:ablation_flow}, we further change the motion reconstruction target from frame difference to RAFT flow~\cite{teed2020raft} and insert the proposed  motion autodecoder into TRX~\cite{TRX}. The results indicate the applicability of the proposed MoLo.

\begin{table}[t]
\centering
\small
\tablestyle{6pt}{1.0}
\setlength{
    \tabcolsep}{
    2.3mm}{
\begin{tabular}
{l|cc|cc}
\hline
			
\hspace{-0.8mm}  \multirow{2}{*}{Setting} \hspace{0.5mm} & \multicolumn{2}{c|}{{SSv2-Full}}  & \multicolumn{2}{c}{{Kinetics}} \\

& \multicolumn{1}{l}{1-shot} & 5-shot  
& \multicolumn{1}{l}{1-shot} & 5-shot    \\ \shline

\hspace{-1mm} Frame Difference & 56.6 & 70.6
& 74.0 & 85.6     \\ 
\hspace{-1mm} RAFT Flow~\cite{teed2020raft}  & \textbf{56.8} & \textbf{71.1}
& \textbf{74.4} & \textbf{85.9}   \\ 
\shline
\hspace{-1mm} TRX~\cite{TRX} & 42.0 & 64.6 & 63.6 & 85.9     \\ 
\hspace{-1mm} TRX + Motion autodecoder  & \textbf{45.6} &\textbf{ 66.1} & \textbf{64.8} & \textbf{86.3}   \\ 
\hline
\end{tabular}
}
\vspace{-3.5mm}
\caption{
Comparison of different motion reconstruction targets and incorporating motion autodecoder into existing TRX.
}
\label{tab:ablation_flow}
\vspace{-3mm}
\end{table}

\begin{figure}[t]
  \centering
   \includegraphics[width=0.99\linewidth]{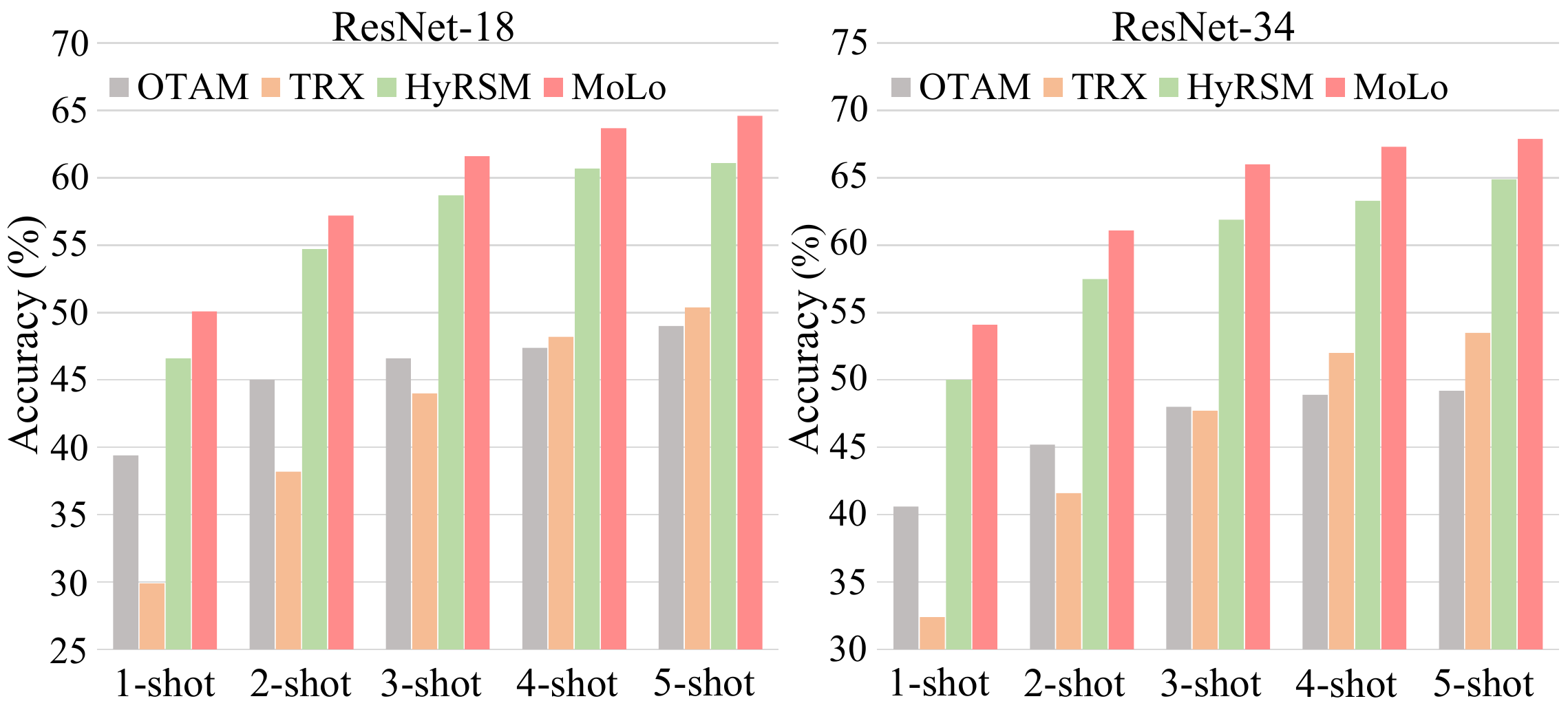}
\vspace{-2mm}
   \caption{Performance comparison of varying backbone depth on the SSv2-Full dataset under the 5-way $K$-shot setting.
   The experiments are carried out with the shot changing from 1 to 5.
   %
   %
   }
   \label{fig:resnet18_renset34}
\end{figure}

\begin{figure}[t]
  \centering
   \includegraphics[width=0.98\linewidth]{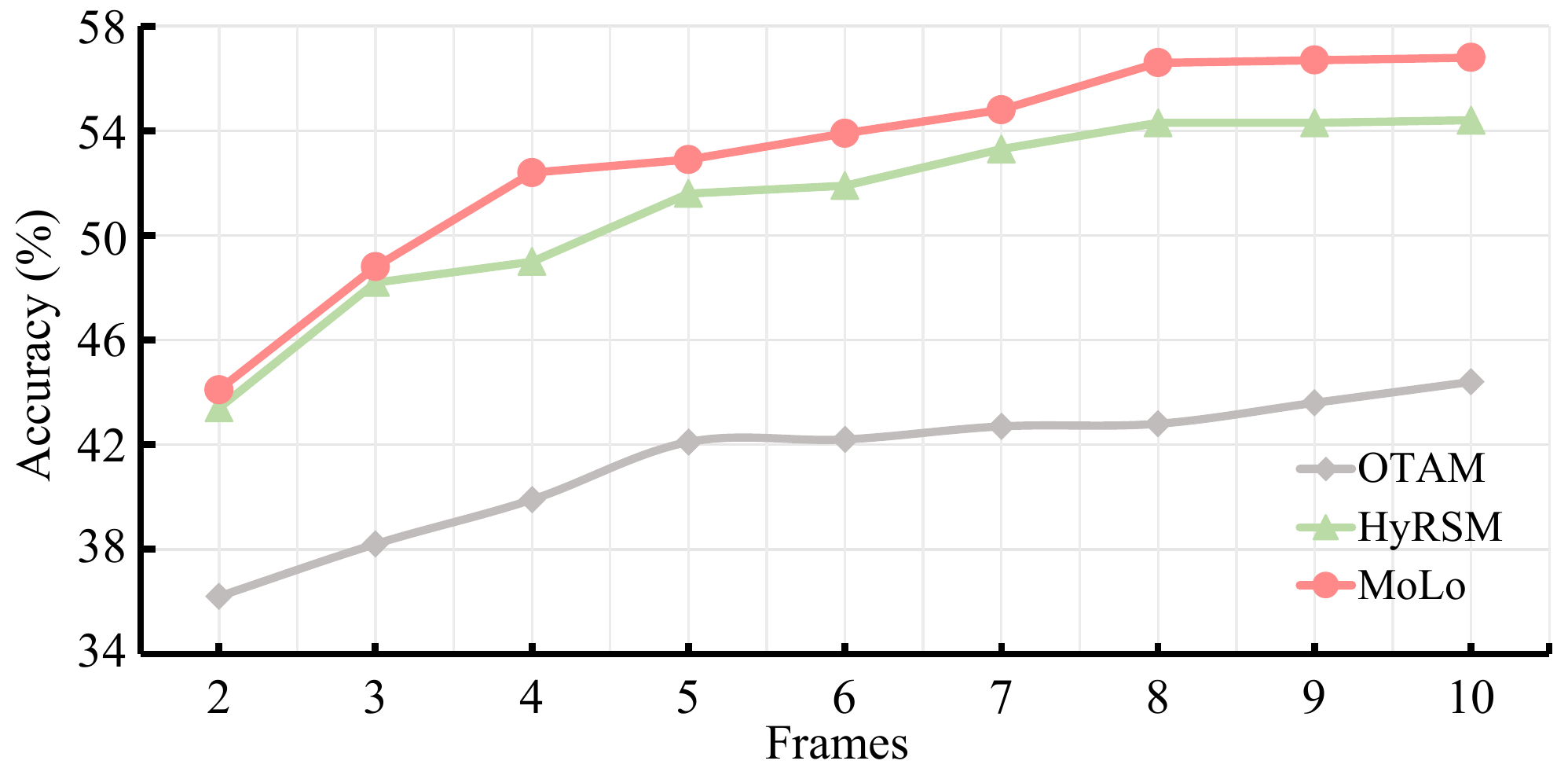}
\vspace{-3mm}
   \caption{
   Ablation study on the effect of changing the number of input video frames under the 5-way 1-shot SSv2-Full setting.
   }
   \label{fig:frame_ablation}
   \vspace{-3mm}
\end{figure}

%

\vspace{1mm}
\noindent \textbf{Varying backbone depth.} 
The previous comparison results are all based on ResNet-50.
To investigate the impact of backbone depth on few-shot performance, we also conduct detailed comparison experiments on ResNet-18 and ResNet-34, as shown in Figure~\ref{fig:resnet18_renset34}.
It can be seen that with the deepening of the backbone, the capacity of the model becomes more extensive, and the performance is also improved.
The proposed MoLo consistently outperforms existing state-of-the-art techniques, including OTAM, TRX, and HyRSM, revealing the scalability of our approach.

\vspace{1mm}
\noindent \textbf{Effect of the number of input frames.} 
For a fair comparison, our MoLo is compared with the current methods under the condition of 8 frames of input. 
To analyze the impact of the number of input frames on few-shot performance, we conduct experiments with input video sampling from 2 to 10 frames.
As shown in Figure~\ref{fig:frame_ablation}, the performance starts to rise and gradually saturate with the increase in the number of input frames.
Notably, our MoLo consistently maintains leading performance, even outperforming the results of OTAM with 8 frames of input when only 2 frames are used, indicating the robustness of our method.

\vspace{1mm}
\noindent \textbf{Different number of temporal Transformer layers.} 
In our framework, the temporal Transformer is adopted to model the temporal long-short associations, and we further explore the impact of the number of Transformer layers on performance. 
The results are displayed in Table~\ref{tab:ablation_transformer}, and we notice that the best results are achieved on the Kinetics dataset when one Transformer layer is applied, and overfitting starts to appear as the number of layers increases.
On the SSv2-Full dataset, the conclusion is slightly different from the above. 
We attribute this to the temporal dependencies being more complex on this benchmark, and more Transformer layers may be required for discovering comprehensive relations on 5-shot.
For the sake of accuracy and model complexity, we utilize one temporal Transformer layer for temporal relationship modeling in our MoLo.
\begin{table}[t]
\centering
\small
\tablestyle{6pt}{1.0}
\begin{tabular}
%
{l|cc|cc}
\hline
			
\hspace{-1.8mm}  \multirow{2}{*}{Setting} \hspace{0.5mm} & 
\multicolumn{2}{c|}{{SSv2-Full}}  & \multicolumn{2}{c}{{Kinetics}}  \\

& \multicolumn{1}{l}{1-shot} & 5-shot  
& \multicolumn{1}{l}{1-shot} & 5-shot     \\ \shline

\hspace{-2mm} Temporal Transformer$\times 1$ & \textbf{56.6} & 70.6
& \textbf{74.0}  & \textbf{85.6}    \\ 
%
\hspace{-2mm} Temporal Transformer$\times 2$ & 56.4 & \textbf{71.7}
& 72.5  & 84.9   \\ 
\hspace{-2mm} Temporal Transformer$\times 3$ & 56.0 & 71.3
& 71.6  & 84.2    \\ 
\hspace{-2mm} Temporal Transformer$\times 4$ & 55.9 & 69.6
& 71.1  & 83.9    \\ 
\hspace{-2mm} Temporal Transformer$\times 5$ & 55.8 & 69.4
& 70.5  & 83.3   \\ 
\hline
\end{tabular}
\vspace{-2mm}
\caption{Ablation study for different number of temporal Transformer layers on the SSv2-Full and Kinetics datasets.
}
\label{tab:ablation_transformer}
\vspace{-1mm}
\end{table}

%
\begin{table}[t]
\centering
\small
\tablestyle{6pt}{1.0}
\setlength{
    \tabcolsep}{
    0.75mm}{
\begin{tabular}
{l|cc|cc}
\hline
			
\hspace{-0.8mm}  \multirow{2}{*}{Setting} \hspace{0.5mm} & 
\multicolumn{2}{c|}{{SSv2-Full}}  & \multicolumn{2}{c}{{Kinetics}}  \\

& \multicolumn{1}{l}{1-shot} & 5-shot  
& \multicolumn{1}{l}{1-shot} & 5-shot     \\ \shline

\hspace{-1mm} Temporal Transformer-only & 53.2 & 68.1
& 72.7  & 84.6    \\ 
%
\hspace{-1mm} Temporal Transformer w/ TAP  & 54.8 & 69.5
& 73.3  & 85.2   \\ 
\hspace{-1mm} \textbf{Temporal Transformer w/ token (MoLo)}  & \textbf{56.6} & \textbf{70.6}
& \textbf{74.0}  & \textbf{85.6}    \\ 
\hline
\end{tabular}
}
\vspace{-2mm}
\caption{Comparison experiments on the effect of learnable token and other variants on the SSv2-Full and Kinetics datasets.
}
\label{tab:ablation_long-short}
\vspace{-1mm}
\end{table}

%
\begin{table}[t]
\centering
\small
\tablestyle{6pt}{1.0}
\setlength{
    \tabcolsep}{
    2.2mm}{
\begin{tabular}
{l|cc|cc}
\hline
			
\hspace{-0.8mm}  \multirow{2}{*}{Setting} \hspace{0.5mm} & 
\multicolumn{2}{c|}{{SSv2-Full}}  & \multicolumn{2}{c}{{Kinetics}}  \\

& \multicolumn{1}{l}{1-shot} & 5-shot  
& \multicolumn{1}{l}{1-shot} & 5-shot     \\ \shline

\hspace{-1mm} Within-video & 54.8 & 70.1
& 73.9  & 85.5    \\ 
\hspace{-1mm} \textbf{Cross-video (MoLo)}  & \textbf{56.6} & \textbf{70.6}
& \textbf{74.0}  & \textbf{85.6}   \\ 

\hline
\end{tabular}
}
\vspace{-2mm}
\caption{Comparison experiments of different long-short contrastive styles on SSv2-Full and Kinetics. ``Within video" means maximizing the representation similarity between local frame features and global features from the same video.
}
\label{tab:ablation_cross_video}
\vspace{-1mm}
\end{table}


%
%
\begin{table*}[ht]
\centering
\small
\tablestyle{6pt}{1.0}
\setlength{
    \tabcolsep}{
    2.6mm}{
\begin{tabular}
{l|cccccc|cccccc}
\hline
			
\hspace{-0.5mm}  \multirow{2}{*}{Method} \hspace{2mm} & 
\multicolumn{6}{c|}{{SSv2-Full}}  & \multicolumn{6}{c}{{Kinetics}}  \\

& \multicolumn{1}{l}{5-way} & 6-way  & 7-way  & 8-way   & 9-way & 10-way 
& \multicolumn{1}{l}{5-way} & 6-way  & 7-way  & 8-way   & 9-way & 10-way   \\ \shline

OTAM~\cite{OTAM}    & 42.8 
& 38.6  & 35.1  & 32.3   & 30.0   & 28.2   
& 72.2   & 68.7  & 66.0  & 63.0  &  61.9 & 59.0 \\
TRX~\cite{TRX}  & 42.0&      41.5    & 36.1  &  33.6   &  32.0   &  30.3   &  63.6   & 59.4  & 56.7  & 54.6  & 53.2  & 51.1  \\  
HyRSM~\cite{HyRSM}    & \underline{54.3} & \underline{50.1}
& \underline{45.8}  & \underline{44.3}  & \underline{42.1}   & \underline{40.0}   & \underline{73.7}   
& \underline{69.5}   & \underline{66.6}  & \underline{65.5}  & \underline{63.4}  & \underline{61.0}  \\
\rowcolor{Gray}
\textbf{MoLo}  & \textbf{56.6} & \textbf{51.6}
& \textbf{48.1}  & \textbf{44.8}  & \textbf{42.5}   & \textbf{40.3}   & \textbf{74.0}  
&  \textbf{69.7}   & \textbf{67.4}  & \textbf{65.8}  &  \textbf{63.5}  &  \textbf{61.3}  \\ 
\hline
\end{tabular}
}
\vspace{-3mm}
\caption{$N$-way 1-shot classification accuracy comparison with recent few-shot action recognition methods on the test sets of SSv2-Full and Kinetics datasets. 
The experimental results are reported as the way increases from 5 to 10.
%
}
\label{tab:compare_Nway}
\vspace{-3mm}
\end{table*}

\vspace{1mm}
\noindent \textbf{Analysis of long-short contrastive objective.} 
We
conduct ablation experiments to show the superiority of
our proposed long-short contrastive objective in Table~\ref{tab:ablation_long-short}.
%
Among the comparison methods, ``Temporal Transformer-only" indicates that the frame features after spatial global average-pooling are directly input into the temporal Transformer and then perform local matching of query-support pairs without long-short contrastive loss.
``Temporal Transformer w/ TAP" means that when calculating long-short contrastive loss, we replace the global token features with the global temporal average-pooling features.
Note that the above variants also contain the motion autodecoder and two few-shot classification heads.
Experimental results show that leveraging the temporal Transformer with a learnable token provides superior performance.  
This suggests the necessity of enforcing the local frame feature to predict the global temporal context during the matching process and the rationality of utilizing the learnable token to adaptively aggregate global temporal features.
In addition, we also explore the effect of the long-short contrastive objective within or between videos in Table~\ref{tab:ablation_cross_video}, and the comparison results demonstrate that the cross-video manner gives more benefits than the within-video form~\cite{wang2022long,dave2022tclr}.
This can be attributed to the fact that the cross-video manner facilitates obtaining a more compact intra-class feature distribution.

\begin{figure}[t]
  \centering
   \includegraphics[width=0.99\linewidth]{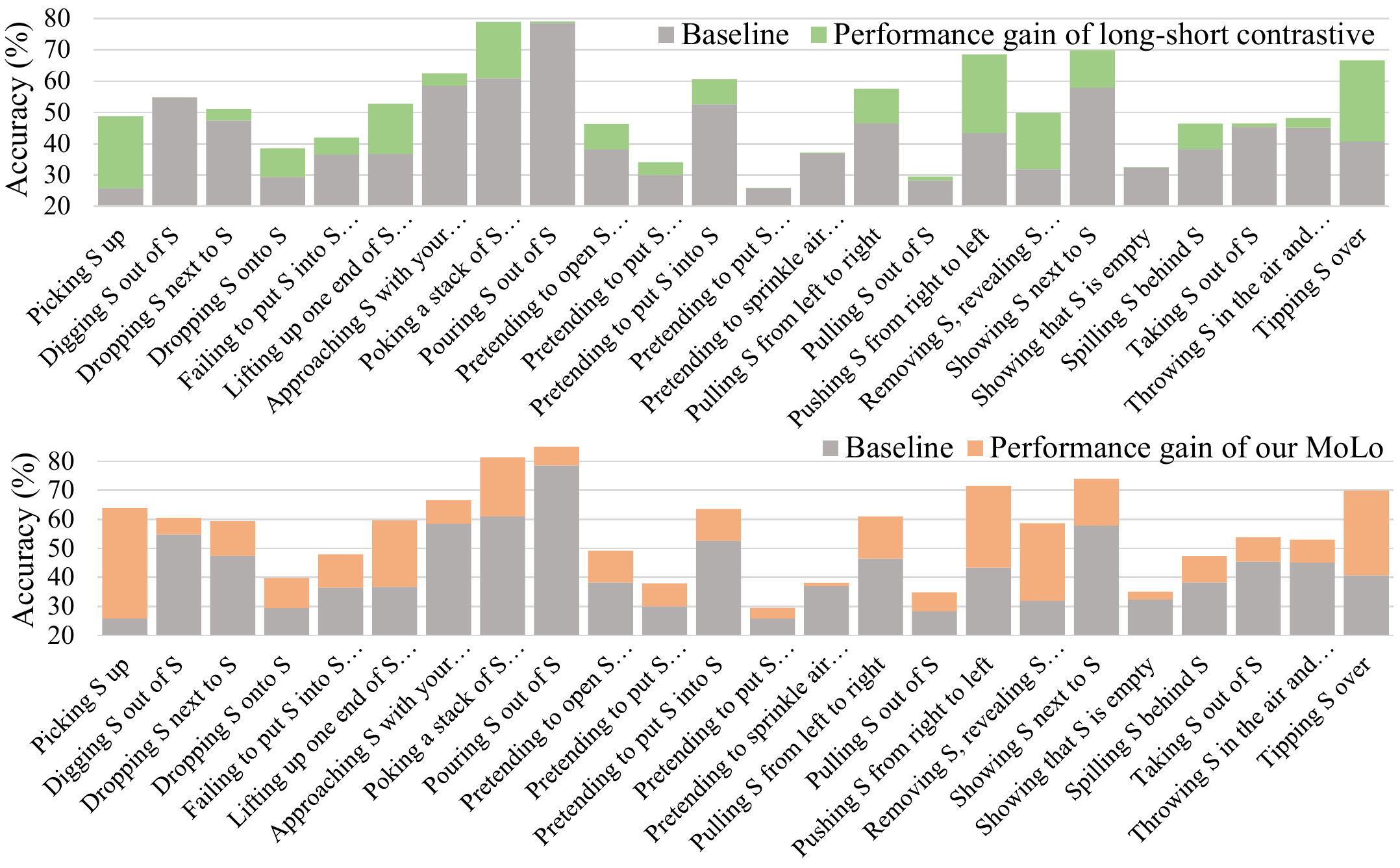}
\vspace{-4mm}
   \caption{Quantitative analysis of 5-way 1-shot class improvements on the SSv2-Full dataset.
   Top: class performance gain of adding the long-short contrastive objective to the baseline method; Bottom: class performance gain of the proposed MoLo compared to the baseline.
   ``S" is the abbreviation of ``something."
   }
   \label{fig:Class_gain}
   \vspace{-2mm}
\end{figure}

\vspace{1mm}
\noindent \textbf{$N$-way few-shot classification.} 
We further conduct ablation
experiments to show the $N$-way 1-shot accuracy, where $N$ ranges from 5 to 10.
In Table~\ref{tab:compare_Nway}, we compare MoLo with existing methods, including OTAM, TRX, and HyRSM. 
%
It can be observed that the larger $N$ leads to greater difficulty in classification, and the performance decreases.
For example, the  10-way 1-shot accuracy of MoLo  drops by 14.3\% compared to the 5-way 1-shot result on SSv2-Full (40.0\% vs. 54.3\%).
Notably, our method consistently achieves superior results under various settings, demonstrating the generalizability of the proposed framework.

\subsection{Visualization analysis}
%
To analyze the impact of the proposed method on the improvements of specific action classes, we perform statistics on the category performance gains compared to the baseline Bi-MHM~\cite{HyRSM} in Figure~\ref{fig:Class_gain}.
From the results, we can observe that by adding our proposed module, each action category has a certain performance improvement. 
Notably, after adding the long-short contrastive objective, both ``Picking something up"  and ``Removing something, revealing something behind" improve by more than 15.0\%.
Similar findings can also be observed in ``Pushing something from right to left" and ``Tipping something over."
This is consistent with our motivation and fully demonstrates that our method can improve the robustness of local frame matching by enforcing long-range temporal perception.
In addition, the performance of each class can be further improved by introducing the motion autodecoder.

\subsection{Limitations}
The 5-shot results of MoLo on some datasets are not satisfactory, and there are marginal gains on several abstract action classes, \eg, ``Showing that something is empty" and ``Pretending to sprinkle air onto something." In the future, we will design a more general method to solve these issues.

\section{Conclusion}
\label{sec:conclusion}
In this paper, we propose a novel MoLo method for few-shot action recognition that consists of a long-short contrastive objective and a motion autodecoder.
The long-short contrastive objective encourages local frame features to predict the global context by contrasting representations. 
%
%
The motion autodecoder is leveraged to recover pixel motions for explicit motion information extraction.
In this way, our MoLo enables robust and comprehensive few-shot matching.
%
%
Extensive experiments
on five commonly used benchmarks verify the effectiveness of our method and demonstrate that MoLo achieves state-of-the-art performance.

\noindent{\textbf{Acknowledgements.}}
This work is supported by the National Natural Science Foundation
of China under grant U22B2053 and Alibaba Group through Alibaba Research Intern Program.


{\small
\bibliographystyle{ieee_fullname}
\bibliography{egbib}
}

\end{document}